\documentclass[11pt]{article}

\usepackage[final]{acl}
\usepackage{times}
\usepackage{latexsym}

\usepackage[T1]{fontenc}

\usepackage[utf8]{inputenc}

\usepackage{microtype}

\usepackage{inconsolata}

\usepackage{graphicx}
\usepackage{booktabs}
\usepackage{multirow}

\def\ie{\textit{i.e.}}
\def\eg{\textit{e.g.}}

\usepackage{caption}
\usepackage{subcaption}
\usepackage{xspace}

\newcommand{\llm}{\texttt{LLM}\xspace}
\newcommand{\api}{\texttt{API}\xspace}
\newcommand{\dfs}{\textit{DFS}\xspace}
\newcommand{\updater}{\textit{updater}\xspace}
\newcommand{\inst}{\texttt{inst}\xspace}
\newcommand{\dataset}{\mathcal{D}}
\newcommand{\workflow}{\mathcal{W}}
\newcommand{\triplet}{(q, \workflow, r)}

\newcommand{\LLM}[1]{\texttt{LLM}\left(#1\right)}
\def\systemname{\textit{ToolGrad}\xspace}
\def\datasetname{\textit{ToolGrad-500}\xspace}

\newif\ifCOMMENTS
\COMMENTStrue
\ifCOMMENTS

\newcommand\zx[1]{{\color{blue} \textbf{ZX: }{#1} }}

\else

\newcommand{\zx}[2]{}

\fi

\usepackage{multirow}
\usepackage{colortbl} 

\usepackage{amsmath}
\DeclareMathOperator*{\argmax}{arg\,max}

\usepackage{threeparttable}

\usepackage{cleveref}
\crefformat{section}{\S#2#1#3} %
\crefformat{subsection}{\S#2#1#3}
\crefformat{subsubsection}{\S#2#1#3}

\usepackage{mathtools}

\usepackage[normalem]{ulem}

\usepackage{xcolor}
\definecolor{mytablegrey}{HTML}{C0C0C0}

\usepackage{listings}

\lstdefinestyle{prompt}{
  basicstyle        = \ttfamily\small,
  backgroundcolor   = \color{gray!10},
  frame             = single,
  rulecolor         = \color{black!60},
  framesep          = 6pt,
  framerule         = 0.6pt,
  breaklines        = true,
  breakautoindent   = false,   %
  breakindent       = 0pt,
  showstringspaces  = false,
  columns           = flexible,
  postbreak         = {},
  xleftmargin       = 0pt,
  xrightmargin      = 0pt,
  moredelim=[s][\color{purple}]{\{}{\}},
}

\renewenvironment{quote}[1][0.04\linewidth]
  {\list{}{\leftmargin=#1\rightmargin=#1}\item\relax}{\endlist}
\newcommand{\myquote}[1]
{
\begin{quote}
\textit{``#1''}
\end{quote}
}

\input{contents/00_teaser}

\title{ToolGrad: Efficient Tool-use Dataset Generation with Textual “Gradients”}

\author{Zhongyi Zhou$^{1,2,3}$, 
Kohei Uehara$^2$,
Haoyu Zhang$^2$,
Jingtao Zhou$^1$,
Lin Gu$^{3,4}$, \\
\textbf{
Ruofei Du$^1$, Zheng Xu$^1$,
Tatsuya Harada$^{2,3}$}\\
$^{1}$Google,
$^{2}$The University of Tokyo,
$^{3}$RIKEN AIP, 
$^{4}$Tohoku University\\
\texttt{zhongyi.zhou.work@gmail.com}, \texttt{harada@mi.t.u-tokyo.ac.jp}
}

\begin{document}
\makeatletter\acl@finalcopytrue
\maketitle
\begin{abstract}
Prior work synthesizes tool-use LLM datasets by first generating a user query, followed by complex tool-use annotations like depth-first search (DFS).
This leads to inevitable annotation failures and low efficiency in data generation.
We introduce \systemname, an agentic framework that inverts this paradigm.
\systemname first constructs valid tool-use chains through an iterative process guided by textual ``gradients'', and then synthesizes corresponding user queries.
This ``answer-first'' approach led to \datasetname, a dataset generated with more complex tool use, lower cost, and almost $100\%$ pass rate.
Experiments show that \systemname models outperform those trained on expensive baseline datasets and proprietary LLMs.
The ToolGrad source code, dataset, and models are available at \url{https://github.com/zhongyi-zhou/toolgrad}.

\end{abstract}

\section{Introduction}
\label{sec: intro}

Tool use empowers large language models (LLMs) by interfacing a parametric model with the external world through API calls.
For instance, RAG~\cite{lewis2020retrieval}, an exemplary tool-use system, demonstrated its impact in reducing LLM hallucination and increasing AI response quality~\cite{shuster2021retrievalaugmentationreduceshallucination}.
Further studies have extended the concept and use programs and database retrieval to enhance LLMs' math reasoning and fact-checking capabilities~\cite{gao2023pal,augenstein2024factuality}.

In practice, teaching LLM to use tools is nontrivial -- its main challenge lies in the dataset.
While prior work has collected large-scale API databases~\cite{shen2023hugginggpt, berkeley-function-calling-leaderboard}, we still lack a scalable method to create a pair of ``user prompt'' and ``tool-use chain'' for training.
Since it is impractical to ask for human annotation at scale, prior work primarily used an agent to search a tool-use path with trial and error.
\autoref{fig:teaser} (top) shows a representative annotation approach, which includes two steps: 1) generate a hypothetical user instruction from a sampled API pool, and 2) use a DFS agent to find its tool-use solution.
This approach is inherently \textit{inefficient} because its core concept is to distill valuable trajectories from a complex agent exploration for training an LLM.
This implies that exploration must be expensive by nature.
More importantly, the exploration has no guarantee of annotation success, causing a waste of agent resources.
As a result, such a tool-use dataset generation usually suffers from 1) a high agent cost and 2) a low pass rate.

To address this issue, this work explores an alternative solution paradigm, \ie, we first generate a ground-truth tool-use chain and then annotate its corresponding user prompt.
Intuitively, an explicit tool-use solution provides more unambiguous information than a prompt, making the annotation, from tool usage to the use query, much easier and requiring only one LLM step.
At the same time, this new problem formulation introduces a new challenge: how can we effectively generate tool-use chains directly from a large-scale API database?

In this work, we introduce ToolGrad, an agentic framework to chain APIs iteratively with mini-batches in a large database.
Inspired by a standard ML optimization loop and TextGrad~\cite{yuksekgonul2024textgradautomaticdifferentiationtext}, we design \systemname to boost textual ``gradients'' by chaining the best API in each iteration of the framework (\autoref{tab: analogy}), 
This is achieved by four modules that perform API proposal, execution, selection, and workflow update, respectively, which resemble the forward inference and backward propagation processes in ML.
Using the framework, we created \datasetname, a tool-use dataset that contains 0.5k samples of user prompts with their corresponding tool calls and AI responses to the user.
Compared to a baseline dataset, ToolBench~\cite{qin2023toolllmfacilitatinglargelanguage}, \datasetname features more complex tool-use data and was generated with lower cost and an almost $100\%$ pass rate.
We further demonstrate that small LLMs fine-tuned on \datasetname can outperform or match SoTA proprietary LLMs, even on out-of-distribution (OOD) datasets with unseen tools.

In summary, this work contributes 1) \systemname, an agentic framework for efficient data generation, 2) \datasetname, a tool-use dataset, and 3) the corresponding fine-tuned models, all of which will be open-sourced to support future research.

\section{Related Work}

\subsection{Tool-use LLMs}

Researchers have studied tool-use LLMs in various fields~\cite{patil2023gorillalargelanguagemodel, huang2024metatool}.
In NLP, tool-use LLMs have shown improved performance in QA~\cite{zhuang2023toolqa}, fact checking~\cite{nakano2022webgptbrowserassistedquestionansweringhuman, augenstein2024factuality, peng2023check} and mathematical reasoning~\cite{gao2023pal, das2024mathsenseitoolaugmentedlargelanguage, schick2023toolformer}.
The impact of tool-use LLMs extends beyond NLP, with notable applications in VQA~\cite{gupta2023visprog, Suris2023vipergpt}, human-computer interaction~\cite{delatorre2024llmr, zhou2024instructpipebuildingvisualprogramming}), and graphic modeling~\cite{huang2024blenderalchemy, du2024blenderllmtraininglargelanguage}.

Datasets play a critical role in advancing the tool-use capability of LLMs.
Initial efforts focused on constructing API databases from various resources, such as Hugging Face APIs~\cite{shen2023hugginggpt} and a community platform~\cite{berkeley-function-calling-leaderboard}.
Given the API databases, there are two primary approaches for creating tool-use datasets that connect user prompts with tool-use actions.
The first group of work relies on human annotations~\cite{zhuang2023toolqa, tang2023toolalpacageneralizedtoollearning}, which is often expensive and difficult to scale up.
Therefore, a large portion of work developed synthetic datasets~\cite{yang2023gpt4tools, wu2024toolplanner}.
ToolBench~\cite{qin2023toolllmfacilitatinglargelanguage}, for example, employs LLMs to generate user queries based on an API database and then performs DFS to search its tool-use solution.
$\tau$-bench~\cite{yao2024taubenchbenchmarktoolagentuserinteraction} synthesizes multi-turn user interactions with a multi-agent simulation.
The latest work further showed that synthetic queries may misalign with human queries in the real world, and thus created new postprocessing modules to rewrite generated user queries~\cite{wu2024toolplanner, zhang2024adaptivequeryrewritingaligning}.

This work follows the synthetic data approach and targets the efficiency issue in the data generation process.
As we will show in experiments, \systemname can generate datasets with more complex tool usage with a lower cost and higher pass rate.

\begin{table*}[]
\centering
\begin{tabular}{cccc}
\hline
\rowcolor[HTML]{EFEFEF} 
{\color[HTML]{333333} } & ML                                                             & {\color[HTML]{333333} \textbf{TextGrad}} & {\color[HTML]{333333} \textbf{ToolGrad}} \\ \hline
Model                                  & $f_\theta(x)$                                                   & $f(x; \phi)$: prompted by $\phi$               & $f(x; \dataset)$: fine-tuned on $\dataset$                    \\
Parameter                              & $\theta$: weights                                              & $\phi$: prompt                           & $\dataset = \{q, \mathcal{W}, r\}$: dataset                   \\
Batches                                & $\{(x, y)\}$: (query, reply)                                   & $\{(x, y)\}$: (query, reply)             & $\{ \api \}$: a small API set               \\
``Gradients''                           & $\nabla_\theta \mathcal{L}\bigl(f_\theta(x), y\bigr)$              & $\llm$ (``criticize it'')                    & $\llm$ (``select the best API'')             \\
Optimizer                              & $\theta_{t+1} \leftarrow \theta_t - \eta\,\nabla_\theta \mathcal{L}$ & LLM updater                              & $\mathcal{W}_{t+1} \leftarrow \mathcal{W}_t.\texttt{add}(\api_t)$      \\ \hline
\end{tabular}
\caption{An analogy of ToolGrad to conventional machine learning and TextGrad~\cite{yuksekgonul2024textgradautomaticdifferentiationtext}. $\dataset$ is a tool-use LLM dataset, composed of many triplets of (query, API workflow, response), \ie, $(q, \workflow, r)$.
}
\label{tab: analogy}
\end{table*}

\subsection{Multi-agent Data Optimization}

LLMs demonstrated their ability to solve problems via simple prompts.
This inspired researchers to create multi-agent collaborative systems for various applications~\cite{park2023generative, wang2023describe}.
For example, AgentCoder~\cite{huang2023agentcoder} improves LLM code generation by having a code generator and a verifier work collaboratively.
MetaGPT~\cite{hong2024metagptmetaprogrammingmultiagent} further simulates human collaboration in software development by simulating different roles like code writers and planners.
Additionally, research shows that agents can self-improve by step-by-step self-criticism~\cite{madaan2023self}.
Copper~\cite{bo2024reflective} further formulates the self-refinement problem with RL, and trains an agent that performs better refinement.

Recent studies formulate LLM agents as operators in classical algorithms for data optimization in various downstream applications~\cite{chen2025can, zhuge2024gptswarm}.
For example, ProTeGi~\cite{pryzant2023automaticpromptoptimizationgradient} optimizes a prompt via LLM-based beam search, which iteratively evaluates, criticizes, and updates an initial prompt design.
TextGrad further defined a unified framework for prompt optimization with textual ``gradients'', and demonstrated its application in a larger domain~\cite{yuksekgonul2024textgradautomaticdifferentiationtext}.

We extend the concept of TextGrad~\cite{yuksekgonul2024textgradautomaticdifferentiationtext} into tool-use LLM dataset generation.
Unlike TextGrad, which optimizes LLMs with better prompts, we aim to generate better datasets to teach LLMs tool usage.

\section{Background: Prompt Optimization with Textual ``Gradients''}
We first review how prior work defines textual ``gradients'' for prompt engineering in an agentic framework.
Note that textual ``gradients'' are not actual mathematical gradients for numerically optimizing objective functions in ML.
Recent work~\cite{yuksekgonul2024textgradautomaticdifferentiationtext} generalizes the mathematical ``gradient'' concept into textual feedback from an LLM critic in an agentic framework, which guides LLMs to update a prompt.

Formally, given an LLM, $f(\cdot; \phi)$, instructed by a prompt $\phi$, prompt optimization aims to iteratively refine an initial prompt $\phi_0$ into an optimized version $\phi_T$, %
so that $\phi_T$ can better instruct $\llm$ for the given downstream task.
This is achieved from an agentic framework with textual ``gradient'' descents that resemble the standard ML optimization. 
In specific, given a batch of downstream task data, $\{(x_i, y_i)\}$, an agentic forward process is defined as $\hat{y_i} = f(x_i; \phi_t)$, where $\hat{y_i}$ is the LLM prediction on a given input $x_i$ using prompt $\phi_t$ on the $t_{\textrm{th}}$ iteration.
The loss signal for the ``gradient'' descent, $\mathcal{L}$, is computed by an LLM agent that criticizes the prediction $\hat{y_i}$.
For example, in article summarization, a critic may comment that a generated summary does not fully summarize the core concept for some reason.
This results in some textual feedback on the summarization tasks, \ie, the textual ``gradients''.
Lastly, another LLM agent edits the prompt conditioned on the critic's feedback, \ie, $\phi_{t+1} \leftarrow \LLM{\phi_t, \mathcal{L}}$.

\begin{figure*}[t]
    \centering
    \includegraphics[width=\linewidth]{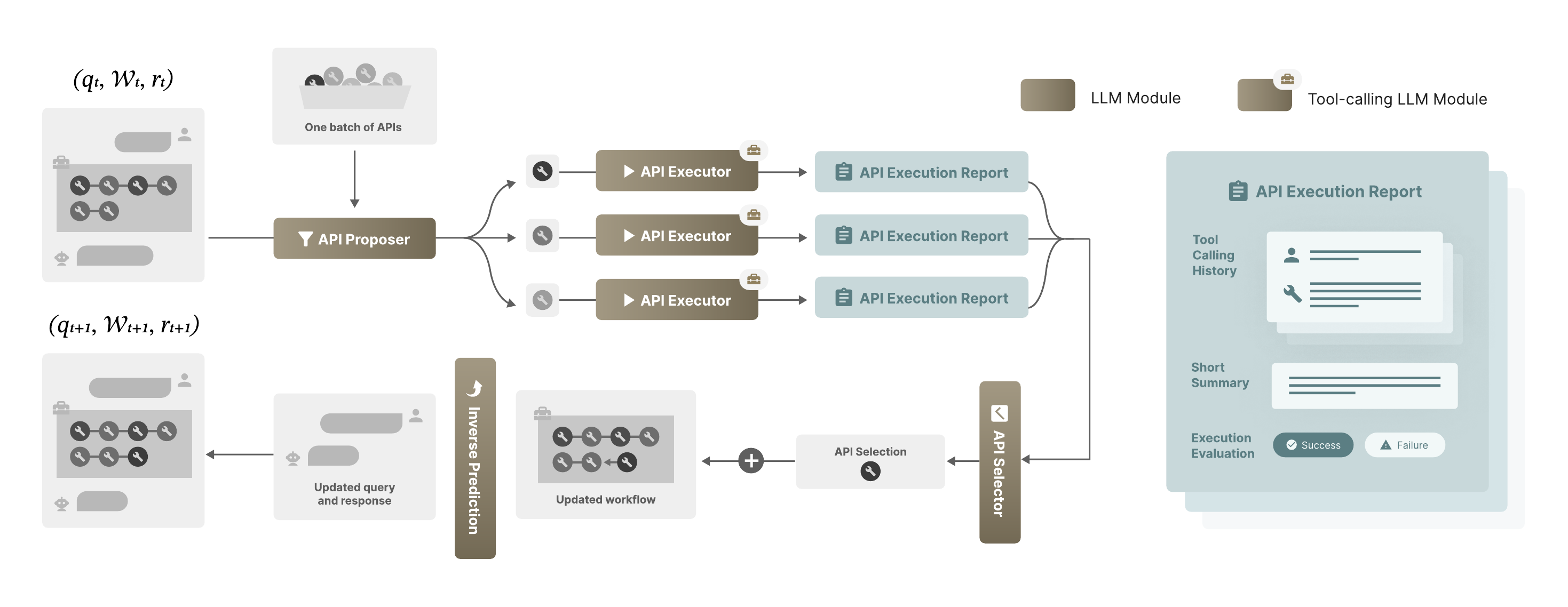}
    \caption{\systemname Framework.
    Each iteration starts with $(q_t, \mathcal{W}_t, r_t)$ and a mini‐batch of APIs.
    An API Proposer first predicts up to $m$ APIs, and then $m$ API Executors perform tool calls and return execution reports. An API Selector finds the most valuable API to chain $\mathcal{W}_t \rightarrow \mathcal{W}_{t+1}$. Lastly, an LLM updater is used to predict $q_{t+1}, r_{t+1}$.
    }
    \label{fig:pipeline}
\end{figure*}

\section{\systemname}
Instead of optimizing prompt engineering, \systemname aims to generate a dataset to teach LLMs tool-use capability.
\autoref{tab: analogy} summarizes the analogy and differences of \systemname, compared to TextGrad and ML.
In practice, generating a tool-use dataset is more complicated than prompt refinement.
Simultaneously updating the model and dataset is an intrinsically challenging analogy to bi-level optimization, as the dataset is used to fine-tune a model, \ie, the internal optimization loop.
Therefore, we leverage LLM feedback for the iterative dataset construction without training an LLM on the dataset in each step.
To achieve such LLM feedback, \ie, the textual ``gradients'', we devise four modules that resemble forward and backward propagation in each step.

\subsection{Tool-use LLMs: Preliminary}

We aim to generate a $\dataset = \left\{\triplet \right\}$ to finetune a tool-use LLM.
$q$ is a user query; $\mathcal{W}$ is an API workflow consisting of a collection of API-use chains: $\mathcal{W} \coloneqq \{ C_1, C_2, \ldots, C_n \}$; and $r$ is the response to $q$ conditioned on $\mathcal{W}$.
A chain is defined as a sequence of API execution steps, $C \coloneqq \api_1 \rightarrow \cdots \rightarrow \api_n$.
An API execution step contains 1) an API ID, 2) the input of this API request, and 3) the response from this API request.

An inference model trained on our dataset differs from the ReAct-based tool-use paradigm, \ie, the default function calling method defined in the OpenAI SDK.
With this dataset, the model is trained to predict all the tool uses in one shot, while ReAct agents predict one tool use in each LLM step.
See more discussion in Appendix~\ref{appendix: three_frameworks}.

\subsection{\systemname: One Iteration Step}
\label{sec: framework}

\autoref{fig:pipeline} visualizes the pipeline of \systemname in each iteration, which contains four core steps: 1) propose top-k APIs to augment the existing API workflows given a mini-batch of APIs, 2) execute the selected APIs, generating API reports, 3) select the best API to augment the current workflow, and 4) update a workflow with the selected API.

\textbf{API Proposer}. The API proposer, $\llm_{pr}$, takes an API mini-batch as input ( $\{\api\}^{bs}$ with size $bs$) and outputs a list of selected APIs with their corresponding instructions on how to use the API for augmenting the current workflow $\workflow_t$: 
\begin{equation}
    \left\{(\api_i, \inst_i)\right\}_{i=0}^{i < m} = \llm_{pr}\left(\{\api \}^{bs}; \mathcal{W}_t\right)
\end{equation}
Parameter $m$ is pre-specified to control the maximal number of API proposals in each step.
Note that we prompt $\llm_{pr}$ with a simple API configuration, and $\llm_{pr}$ cannot respond with a tool-calling request.
This design distills the most valuable APIs for use in subsequent requests, thereby improving overall system efficiency.
Our intuition is that 1) most of the APIs in a randomly sampled batch are irrelevant to the current workflow, and 2) providing simple API configurations is sufficient for an LLM to decide which APIs are worth further in-depth execution.
Therefore, $m$ must be much smaller than $bs$ to achieve such efficiency in practice.

\textbf{API Executor}. The API proposals are then sent to $m$ API executors, $\{\llm^T_{ex}\}^m$.
$\llm^T$ is denoted as a tool-calling LLM agent that can return tool-calling requests, as opposed to $\llm$, which returns standard responses to the user.
$\llm^T_{ex}$ takes an API proposal ($\api_i$, $\inst_i$) as input and return a report,
\begin{equation}
    \text{rep}_i = \llm^T_{ex}\left(\api_i, \inst_i \right).
\end{equation}
The report contains the following information: 1) a full record of the API request history and 2) a boolean variable showing whether the execution is successful.
This is the most expensive step in the \systemname framework because each selected API is paired with an LLM agent for parallel execution. 
This verifies the necessity of our API proposer step, which performs filtering, in one LLM step, to avoid redundant API calls.

\textbf{API Selector}.
Given a set of execution reports $\{\text{rep}_i\}$, we design an API selector, $\llm_{sel}$, to choose the best API that can augment the current workflow $\workflow_t$. 
\begin{equation}
\label{value_func_llm}
\begin{split}
j &= \argmax_i V \left(\{\text{rep}_i\}^m, \workflow_t \right) \\
  &\sim \llm_{sel} \left(\{\text{rep}_i\}^m, \workflow_t \right),
\end{split}
\end{equation}

where $V(\cdot, \cdot)$ is a hypothetical value function. In practice, instead of defining $V$ and performing $\argmax V(\cdot, \cdot)$, we use an LLM as its proxy.
Intuitively, $\argmax V(\cdot, \cdot)$ is a process that chooses the most valuable API from the reports, and we hypothesize that an LLM can achieve this task conditioned on the API execution reports, $\{\text{rep}_i\}^m$, and the current workflow, $\workflow_t$. 
In addition, we instruct $\llm_{sel}$ to specify which chain $C_k \in \workflow_t$ the selected API ($\api_j$) augments -- or to create a new chain if necessary.
Therefore, the following equation shows the API selector step at \systemname:
\begin{equation}
\begin{split}
    j, k &= \llm_{sel}\left(\{\text{rep}_i\}^m, \workflow_t \right), \\
         &\quad 
         { \footnotesize
         \text{where } 
         \begin{cases}
            j \text{~is the selected \api id for~} \api_j, \\
            k \text{~is the chain id for~} C_k.
         \end{cases}
         }
\end{split}
\end{equation}

The API selection is the core step that performs the ``gradient'' computation in our optimization loop (\autoref{tab: analogy}).
As opposed to the LLM critic step that uses textual feedback as ``gradient'', our API selector chooses a discrete API to augment $\mathcal{W}_t$ as ``gradients'' of data generation in \systemname.

\textbf{Workflow Updater}.
$j$ and $k$ from the API executor provide clear information on 1) which API from the mini-batch the workflow updater should use and 2) where (at which chain) the updater should append the API to.
Therefore, the workflow updating process can be clearly defined as follows without using LLMs.
\begin{equation}
    \workflow_{t+1} \leftarrow \workflow_t.\texttt{add}(\api_j, C_k)
\end{equation}

On the other hand, once $\workflow_t$ is updated to $\workflow_{t+1}$, we should also update $(q_t, r_t)$ to maintain the coherence of the sample triple $\triplet$. 
Therefore, in the workflow updating step, we perform the following LLM step:
\begin{equation}
    q_{t+1}, r_{t+1} = \llm_{\updater}(\workflow_{t+1})
\end{equation}

Intuitively, this step resembles summarization tasks that convert detailed texts (\ie, a tool-use workflow) to ambiguous messages (\ie, a user query and its response).
This inverse prediction process is much more straightforward than the standard forward pass that explores answers with a given user query: $\workflow, r = \llm_{\dfs}(q)$, where $\llm_{\dfs}$ is an agent using DFS~\cite{qin2023toolllmfacilitatinglargelanguage}.

\subsection{Sampling Negative APIs}
Given the $\triplet$ with the ground-truth tool uses, we post-process it by sampling negative tools.
The objective is to simulate a real-world use scenario where an agent can access more APIs than necessary.
Prompting the LLM with every API configuration is impractical given our API database’s size (8k).
Therefore, we aim to simulate an RAG-like use case in our dataset, in which the agent first samples top-$p$ APIs based on the text-embedding similarity and then prompts an LLM with the $p$ APIs only.
Formally, given a ground truth set $\{\mathcal{W}\}^n$ of $n$ positive APIs, we sample the top-$(p-n)$ APIs most similar to these positives as our negative samples.

\subsection{Generation Configuration}
In this work, we choose the number of API proposals as $m=3$, and the API batch size $bs=50$.
Each generation loop takes 10 iteration steps, which we observed is sufficient to generate complex API-use workflows.
We chose $p=10$ when sampling negative APIs and \textit{gemini-2.5-flash-lite} as our LLM for data generation.
We chose the lite model for its cost-effectiveness, and its benefit in low-latency response capability.
The low-latency benefits are crucial for our framework to generate data within a reasonable amount of time.
The framework for 500 times using different seeds, constituting \datasetname.
We then further format the generated samples into a supervised finetuning set for single-turn tool uses, equivalent to the tasks defined in Berkeley Function Calling Leaderboard (BFCL) v1/v2~\cite{patil2023gorillalargelanguagemodel}.
The LLM is provided with an OpenAI-style tool-use definition and trained to output a Python-style tool use.
Appendix~\ref{appendix: dataset} provides more details on our prompt templates.

\section{Experiments}

We conducted experiments to demonstrate that \systemname 1) generates high-quality data with low cost, and 2) leads to a finetuned model with better tool use performance.

\subsection{Setups}

\textbf{Model training and baselines}.
We then finetune three Gemma-3 models (1B, 4B and 12B) on these two datasets using supervised finetuning.
More training configs are documented at Appendix \ref{appendix: train_config}.
We chose baselines from two aspects.
We first chose ToolBench~\cite{qin2023toolllmfacilitatinglargelanguage} as our query-first baseline dataset, and their DFS framework as our query-first data generation framework.
We evaluate the quality of frameworks both explicitly (generation cost and generated data complexity) and implicitly (model performance trained on their generated datasets).
To achieve a fair implicit comparison, we also finetune the Gemma-3 models on ToolBench datasets, and report results based on the finetuned models.
We also consider model-wise baselines, where we experimented with Gemini-2.5, GPT-5 and Claude-4.5 series models as representatives of SoTA proprietary LLMs.
In addition, we chose ToolACE~\cite{liu2025toolacewinningpointsllm} and Hammer~\cite{lin2024hammerrobustfunctioncallingondevice} as baselines that represent finetuned models specifically for tool uses. 

\textbf{Evaluation benchmarks}.
We evaluated the finetuned models on two benchmarks. We first report results on ToolBench-I3~\cite{qin2023toolllmfacilitatinglargelanguage}, the most challenging track in ToolBench with cross-category tool use.
Therefore, we report our model performance on BFCL~\cite{patil2023gorillalargelanguagemodel}.
Note that there is minimal tool definition overlap between ToolBench and BFCL, so this evaluation helps us understand whether our models can successfully use unseen tools. 
Our evaluation focuses on BFCL-v1/v2, the single-turn tool-use track, because multi-turn tool use and agent use (the latest v3/v4 track) are out of scope for our generation framework.
We leave it as future work to explore enhancing agent performance with the \systemname-generated dataset.

\textbf{Metrics}. We report two groups of metrics to evaluate a given dataset generation approach.
To evaluate the \textit{cost} of data generation, we report 1) the pass rate, 2) the number of ground-truth tool uses, which measures the complexity of generated data, 3) the number of LLM/tool calls, and 4) the LLM API cost.
We also report the \textit{performance} of models finetuned on the given dataset on ToolBench and BFCL. We reused their corresponding evaluation metrics. Note that on ToolBench, we use absolute LLM judge scores instead of the original win rate metric. Our absolute scoring system fits more on our experiments with multiple baselines.
Appendix~\ref{appendix:llm-judge} provides more details on our design of the LLM judge, including the prompt design.

\begin{table}[t]
\centering
\begin{tabular}{lcc}
\toprule %
& DFS & \systemname         \\ 
\midrule %
Pass rate ($\%$)~$\uparrow$      & 63.8  & \textbf{99.8}      \\
\# of gt tool uses~$\uparrow$    & 2.1 & \textbf{3.4}        \\ 
LLM cost~$\downarrow$            & 64.5  & \textbf{63.9}       \\
Tool cost~$\downarrow$           & 34.3  & \textbf{20.0} \\ %
\bottomrule %
\end{tabular}
\caption{Generation efficiency comparison between DFS~\cite{qin2023toolllmfacilitatinglargelanguage} and \systemname.}
\label{tab:data_gen_efficiency} %
\end{table}

\begin{table*}[t]
\centering
\begin{tabular}{c|ccc|ccc|ccc|ccc}
\toprule
\multirow{2}{*}{Model} & \multicolumn{3}{c}{ToolGrad} & \multicolumn{3}{c}{Gemini 2.5} & \multicolumn{3}{c}{Claude 4.5} & \multicolumn{3}{c}{GPT-5} \\
                       & 1B       & 4B      & 12B      & lite     & flash     & pro      & haiku    & sonnet    & opus     & nano    & mini   & base   \\ \midrule
Score                  & 14.1     & \underline{17.6}    & \textbf{19.6}     & 6.9      & 8.5      & 11.4     & 12.8     & 13.5      & 15.4     & 15.4    & 14.7   & 12.7   \\ \bottomrule
\end{tabular}
\caption{ToolBench single-turn tool-use evaluation results. We highlight the \textbf{best} and \underline{second best} value.}
\label{tab: toolbench_single_results}
\end{table*}

\begin{table*}[]
\centering

\begin{tabular}{lcccccccccc}
\toprule
\multirow{2}{*}{} & \multicolumn{3}{c}{Gemma-3-1B} & \multicolumn{3}{c}{Gemma-3-4B} & \multicolumn{3}{c}{Gemma-3-12B} & \multirow{2}{*}{\begin{tabular}[c]{@{}c@{}}Gemini-2.5\\ flash-lite\end{tabular}} \\
  & Base & ToB & ToG & Base & ToB & ToG & Base & ToB & ToG & \\ \hline
Standard & 1 & / & \underline{14.1} & 11.2 & / & \underline{17.6} & 9.8 & / & \textbf{\underline{19.6}} & 6.9 \\
ReAct & 0 & 3.3 & / & 0.1 & \underline{17.6} & / & 0.5 & 18.1 & / & 12.3 \\
DFS & 0 & 3.2 & / & 0.1 & 13.4 & / & 0 & 12.3 & / & 15.6 \\ \bottomrule
\end{tabular}

\caption{Comparison of models trained on ToolGrad and ToolBench datasets on ToolBench. ToB and ToG denote finetuning on ToolBench and ToolGrad datasets, respectively. The best performance per model size is \underline{underlined}, and the \textbf{global best} is bolded.}
\label{tab:toolbench_results}
\end{table*}

\begin{figure*}[]
\centering
\includegraphics[width=\linewidth]{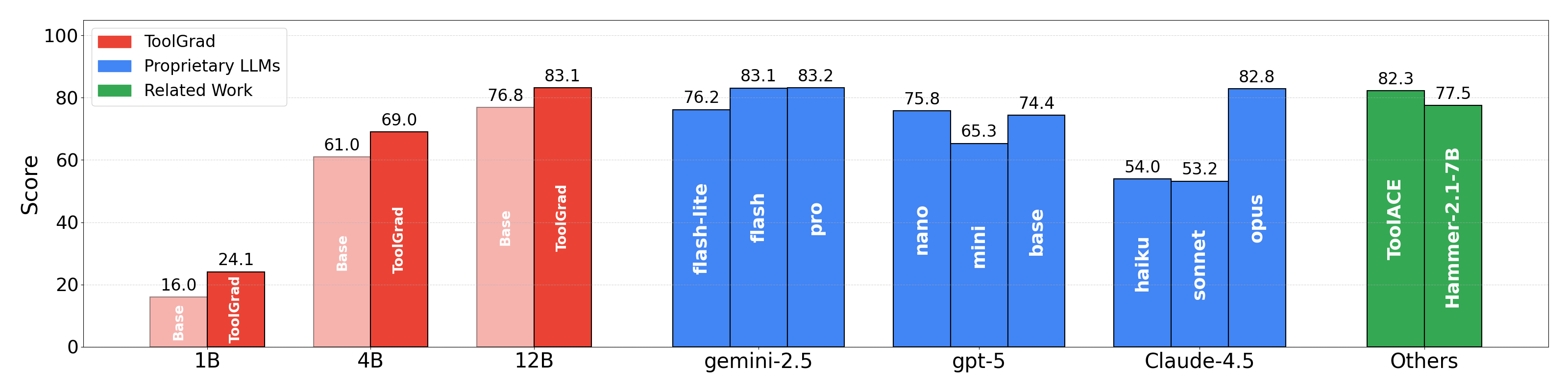}
\caption{BFCL results. The graph shows single-turn results, and the corresponding breakdown into subscales is available at~\autoref{tab: bfcl_subscale_1} and \autoref{tab: bfcl_detailed_subscale}. For those supporting both "prompt" and "function call" setups, we chose the "prompt" condition as a fair comparison with Gemma-3, where there is no official function call support.}
\label{fig:bfcl_results}
\end{figure*}

\subsection{Main Results}

\textbf{Data Efficiency}.
\autoref{tab:data_gen_efficiency} summarizes the cost-performance results comparison between our answer-first data generation framework, ToolGrad, and a DFS-based query-first approach on ToolBench~\cite{qin2023toolllmfacilitatinglargelanguage}. 
ToolGrad achieves 99.8\% pass rate -- a significant improvement from 63.8\% for DFS, while producing more complex chains (an average of 3.4 ground-truth tool uses vs. 2.1 for DFS).
More importantly, \systemname cuts down on the generation cost: LLM invocations drop from 64.5 to 63.9, and tool-use steps fall from 34.3 to 20.0. 
The results demonstrate the high efficiency of \systemname for data generation: it generates more complex tool-use chains with a higher pass rate and lower cost.

We reviewed the failure log and found that the agent failed to receive a successful response from 3 selected APIs across all 10 iterations, and thus saved an empty data sample.
In practice, this case happens extremely rarely (0.2\%).

\textbf{Results on ToolBench}. We then report how effective our ``cheap'' dataset can be used to teach LLM tool usage.  \autoref{tab: toolbench_single_results} compares single-turn tool-use performance between our finetuned small LLMs and SoTA proprietary LLMs.
Results show that our 12B and 4B models ranked first and second, outperforming even our teacher model, ``gemini-2.5-flash-lite''. 

\autoref{tab:toolbench_results} further compares our models on ToolGrad to the same models trained on ToolBench and the base Gemma models, using the ToolBench test set.
We first compared ToolGrad models with base Gemma models, and results show that ToolGrad can effectively enhance Gemma's tool use capability (1B: +13.1, 4B: +6.4, 12B: +9.8). 
We then compare our models with ToolBench-trained models, which are 1) trained on more expensive datasets (see training budget data Appendix~\ref{appendix:training_budgets}), 2) use an advanced inference framework, and 3) evaluated in-distribution with their test set.
Despite the ``unfair'' setup that strongly favors our baselines, the ToolGrad models still outperform the corresponding ToolBench models (with a tie on the 4B model).
Additional discussion on \autoref{tab:toolbench_results} is available in Appendix~\ref{appendix:toolbench_discuss}.

Note that ToolBench evaluation heavily rely on LLM judges, so we conducted a small-scale human evaluation to test its alignment with human scores. We randomly selected 8 queries and asked two human raters to evaluate the answers from the 12 models in \autoref{tab: toolbench_single_results}, yielding 96 ratings per rater.
Further details regarding the study are available in Appendix~\ref{appendix:human-eval}.
Our analysis indicates that the averaged judge scores correlate strongly with human ratings ($\rho= .88, p < .001$).
This observation aligns with existing literature~\cite {chan2023chatevalbetterllmbasedevaluators, wang2025improvingllmasajudgeinferencejudgment} and serves to validate both our judge design and the model performance results on ToolBench.

\textbf{Results on BFCL}.
\autoref{fig:bfcl_results} shows our evaluation results on BFCL.
We only report overall score of single-turn tool uses in the figure.
The score breakdowns are available in \autoref{tab: bfcl_detailed_subscale} and \autoref{tab: bfcl_subscale_1}.

The results demonstrate that our tool-use capability generalizes effectively to unseen tool usage scenarios. Specifically, the 1B, 4B, and 12B models achieve score increases of +8.1, +8.0, and +6.3, respectively. \autoref{tab: bfcl_subscale_1} further confirms these performance gains on both the live (+1.93, +4.74, +4.22) and non-live (+14.19, +11.34, +8.37) subscales. While the trend indicates that our fully synthetic dataset most significantly boosts non-live (synthetic) benchmark scores, it notably enhances model performance on real human data benchmark as well. \autoref{tab: bfcl_detailed_subscale} shows that our ToolGrad dataset is particularly beneficial to boost ``Multiple Parallel'' performance, defined as a combined task of 1) tool choices and 2) parallel tool calls. 
This result well aligns with our results reported in~\autoref{tab:data_gen_efficiency}, where we show the \systemname dataset is good at generating complex tool uses (with low cost).

\autoref{tab: bfcl_subscale_1} also shows that ToolGrad-12B ranks the second in BFCL (-0.1 compared to Gemini 2.5 Pro).
ToolGrad-12B also outperforms ToolACE~\cite{liu2025toolacewinningpointsllm}, trained on a more advanced tool set than ToolBench, and Gemini-2.5-flash-lite, its teaching model in the \systemname framework.

\subsection{Scaling Study}
\label{sec:scaling}
\begin{figure}[t]
\centering
\includegraphics[width=\linewidth]{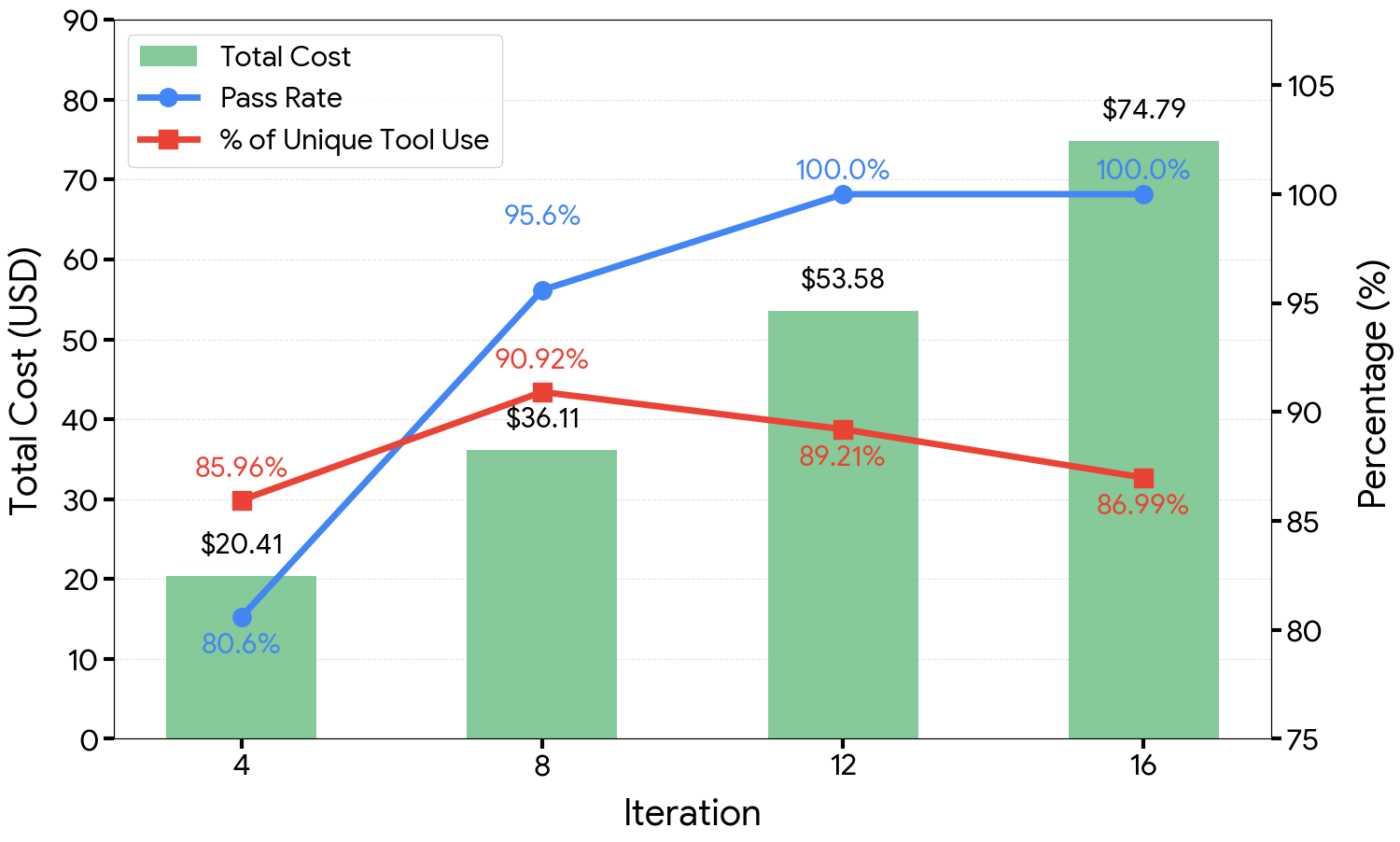}
\caption{Cost-effective analysis on iterations scaling.}
\label{fig:scaling_iteration}
\end{figure}

\begin{figure}[t]
\centering
\includegraphics[width=\linewidth]{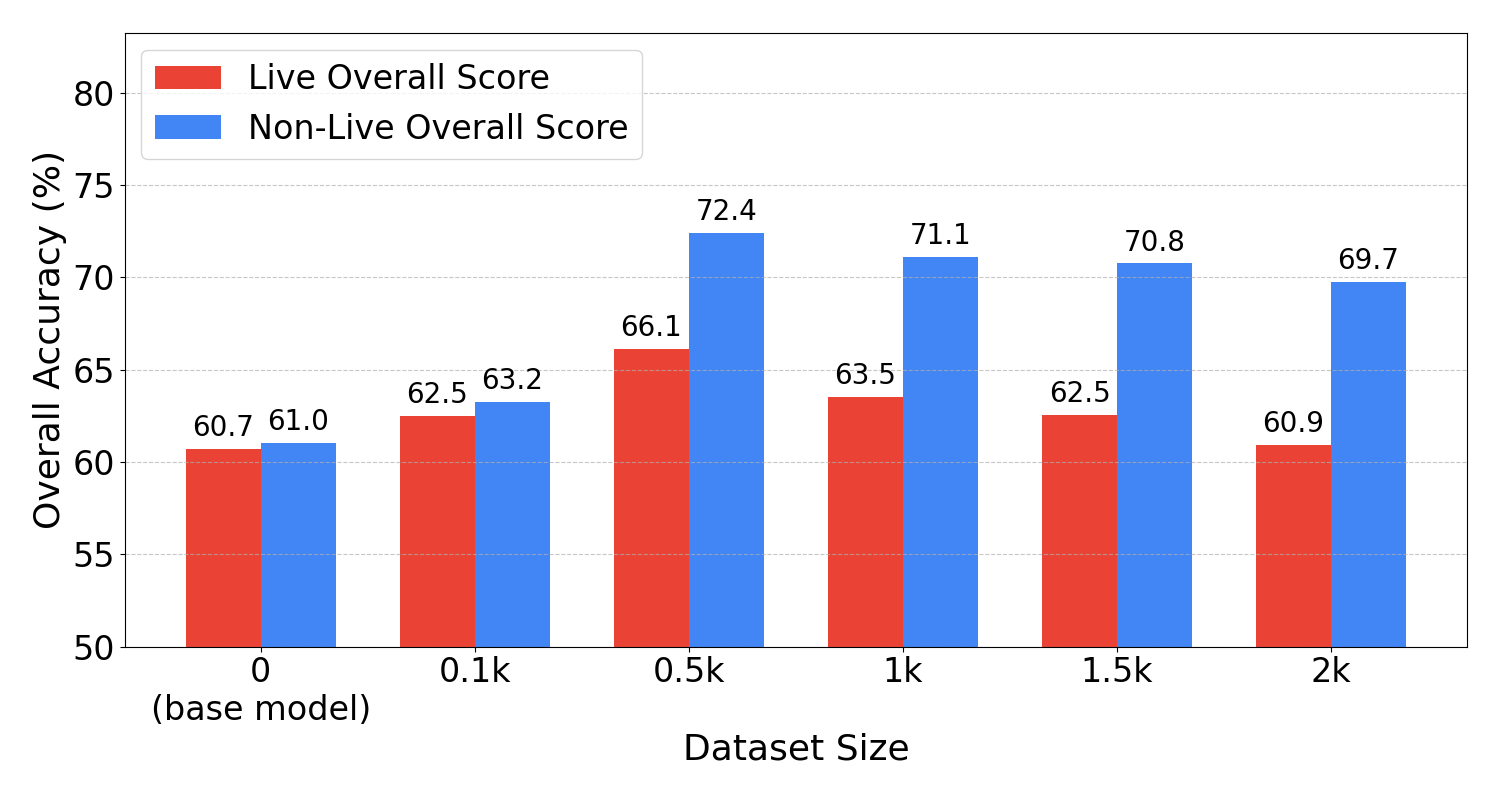}
\caption{Gemma-3-4B model accuracy on BFCL when scaling ToolGrad samples.}
\label{fig:scaling_samples}
\end{figure}

Different from other modularized projects that evaluate their full pipeline with an ablation study, modules in our framework work as a whole and cannot be reasonably ablated.
Here, we conducted a scaling study that explores how \systemname performance varies by scaling the number of iterations and the number of samples.

\textbf{Scaling iterations}.
We chose iteration = 4, 8, 12, 16 and number of samples = 500 for the study, and reported how 1) pass rate, 2) percentage of unique tool use, and 3) LLM API cost for generating the data using \systemname.
\autoref{fig:scaling_iteration} summarizes our results using Gemini 2.5 flash.
The figure shows that the pass rate tends to saturate between iteration = 8 and 12, demonstrating our choice of iteration=10 when crafting \datasetname as a reasonably cost-effective choice.
The ``\% of Unique Tool Use'' curves demonstrate one major challenge in the ``scaling law'', i.e., the agent tends to generate similar tool uses in our training set.
Such repetitive synthetic tool-use data will inevitably harm our model training when we scale our framework for large-scale data generation.

\textbf{Scaling samples}.
We further studied the scaling effect on the number of samples. We chose 10 iterations per generation and trained Gemma-3-4B on 100, 500, 1k, 1.5k, and 2k data, using the same finetuning configuration for ToolGrad-4B. We then evaluate the finetuned models on BFCL.
Results in \autoref{fig:scaling_samples} show that, while all finetuned models still outperform the baseline, the model performance first increases and then decreases with the scaling of the dataset.
These results provide further evidence showing the existence of a saturation point with our current \systemname system design.
We encourage future work to formally investigate this phenomenon and then contribute to breaking this performance saturation point.

\subsection{Discussion}

\textbf{Contaminated Data with Unsolvable Queries}.

\autoref{tab: toolbench_single_results} shows that all models perform poorly in the ToolBench.
After investigation of the ToolBench test query, we find that there exist suboptimal queries where the given tool sets cannot support solving the query.
This issue is inevitable in the query-first data generation framework, and it is almost impossible to guarantee the resolvability of a generated query.
Additionally, even though the query is solved by DFS exploration, the Toolbench training set considers the traces that lead to success as the ``ground truth'' tool-use traces for LLMs to learn.
However, we find such ``ground truth'' tool-use traces contain low-quality tool-use steps like wrong tool use that lead to data contamination.
In comparison, our answer-first generation framework can avoid both issues: 1) all generated data are verified because the generation starts with at least one valid tool use, and 2) our framework does not allow the tool-use failure step to enter our training dataset. 

\textbf{Intelligence Bootstrap with \systemname}.
One common approach to eliminate the data contamination in query-first approaches is to filter out failure samples~\cite{du2024blenderllmtraininglargelanguage}.
This approach is also suboptimal -- It limits an agent's learning space to problems that a teacher agent can solve.
As a result, a student cannot outperform a teacher model (\eg, ToolLlama fails to beat GPT-4~\cite{qin2023toolllmfacilitatinglargelanguage}).
In contrast, our results on both benchmarks provide strong signals of bootstrapping an LLM intelligence with our unique design of answer-first data generation.
For example, \autoref{tab: toolbench_single_results} shows that even the ``1B'' model can outperform its teacher model, ``gemini-2.5-flash-lite''.
More interestingly, Gemma-3-12B already outperforms ``gemini-2.5-flash-lite'' in both benchmarks, but ``gemini-2.5-flash-lite''-generated data can still be useful to effectively teach a better ToolGrad-12B model.

\textbf{Agent Memory}.
In Section~\ref{sec:scaling}, we reported that the model performance fails to increase with the scaling of our data generation. We believe one major reason is that \systemname does not contain a memory mechanism.
That being said, each data sample in \systemname is generated independently. 
This explains the finding under \autoref{fig:scaling_iteration}, where the framework tends to generate similar tool use in the training set.
With a shared memory module, we envision a future version of \systemname will be able to avoid proposing repetitive tool usage during the data generation, and thus create a higher quality tool use training set.

\section{Conclusion}
This work introduces \systemname, a framework for efficient tool-use dataset generation.
Our core concept is to first generate tool-use answers with textual ``gradients'', followed by query generation.
We further contribute a high-quality dataset, \datasetname, generated with a lower cost and almost $100\%$ pass rate.
Experiments show that models trained on \datasetname outperform those trained on expensive baseline datasets and proprietary LLMs.

\section{Limitations}
\textbf{Multi-step tool use with reasoning}. Our models are limited in inference with the ReAct/DFS framework because our fine-tuning dataset lacks reasoning examples.
Recent work has introduced an increasing number of tool-use frameworks~\cite{lu2025octotoolsagenticframeworkextensible}, and it is underexplored how our superior performance can be generalized into a broader range of frameworks.

\textbf{Post-training with reinforcement learning (RL)}. This work focuses on demonstrating the usage of our dataset with supervised fine-tuning (SFT).
Recent exploration highlights the benefit of teaching LLM tool usage with RL~\cite{qian2025toolrlrewardtoollearning}.
The value of the datasets we generate for RL remains underexplored. 

\textbf{Aligning human behaviors in generated queries}. While the ToolGrad framework enhances the generation efficiency of synthetic tool-use datasets, it does not address another critical issue of synthetic datasets: the alignment of real-world human behaviors.
That being said, an LLM-generated user query may not accurately reflect how real-world humans express their intentions when interacting with LLMs.
For example, generated queries may lack linguistic diversity.
Future work should consider post-processing the generated queries~\cite{wu2024toolplanner, zhang2024adaptivequeryrewritingaligning} or take humans in the ToolGrad framework, leading to another analogy to interactive machine learning (IML).
With the combined effort of higher efficiency and better human alignment in synthetic approaches, we believe that future agents will be able to rapidly bootstrap their tool-use capability through self-instruction.

\textbf{Scaling Failure}.
A major benefit of synthetic data is that it can scale up more easily than real human data collection.
However, in Section~\ref{sec:scaling}, we show the scaling plateau appears at a small number, which limits the motivation of utilizing synthetic data instead of real human data for LLM post-training.
We encourage future work to push this scaling limit.

\bibliography{contents/99_reference}
\appendix

\newpage

\clearpage
\appendix
\section*{Appendix}

\section{ToolGrad System}
\subsection{Prompts}

The following content shows the prompt templates used in this work, including those in four LLM modules~\autoref{fig:pipeline} of \systemname (\autoref{lst:api-proposer}, \autoref{lst:api-executor}, \autoref{lst:api-selector}, and \autoref{lst:inverse-prediction}) and the template we used in the LLM judge~\autoref{lst:llm-judge}.

\begin{lstlisting}[style=prompt, caption={``API Proposer'' template.}, label={lst:api-proposer}]
You are tasked with augmenting an API-use workflow with more APIs from a given library so that it can serve for more advanced tasks.
Given the following information that provides the context, please make three API-use proposals to augment the current workflow.

The current workflow:
{workflow_cur}

The following is a pool of APIs that you can use:
{api_all}

CRITICAL - Instruction Format Requirements:
- Each proposal's "instruction" field must be an ACTIONABLE COMMAND that tells an agent executor HOW to use the proposed API(s).
- Instructions must be IMPERATIVE commands (e.g., "Get...", "Use...", "Retrieve..."), NOT descriptive summaries.
- DO NOT reference previous executions or results (e.g., avoid "This tool was used..." or "It returned...").
- The instruction will be given to an agent executor that will use it to decide which tools to call and with what inputs.

Notes:
- Please reply in the required data structure.
- To select an API, you should return its name.
- If you do not have any additional tools to propose, you can respond with None.
\end{lstlisting}

\begin{lstlisting}[style=prompt, caption={``API Executor'' template.}, label={lst:api-executor}]
You are tasked with exploring an API based on a given plan.
The following shows a guide for you to follow:
1. Verify whether the API-calling result provides a reasonable response for the given plan.
2. Report `success = False` if you fail to get a reasonable result, and explain why.
3. Report `success = True` if you get a reasonable result that addresses the plan, and provide justification for the success.
4. If you report `success = True`, you should also report which function calling step leads to the success.

Retry Logic:
- The API may return bad results (e.g., error messages, completely irrelevant data, or responses that don't address the plan).
- In such cases, you should try again with different input parameters if reasonable alternatives exist.
- You have a maximum number of iterations to complete the task. Use them wisely.
- If retries fail or no reasonable alternative inputs exist, report `success = False`.

The following is the plan:
{plan}

Notes:
- Success criterion: If the API execution provides a reasonable result that addresses the given plan, report `success = True`.
- If an API fails to execute or returns unusable results after reasonable attempts, report `success = False`.
\end{lstlisting}

\begin{lstlisting}[style=prompt, caption={``API Selector'' template.}, label={lst:api-selector}]
You are an API selector.

You need to select one API or refuse to select any API from the given list of APIs to augment the current workflow.

The current API-use workflow:
{workflow_cur}

Reports from the proposed APIs:
{api_reports}

When you select an API, you need to make the following decisions:
1. Determine whether any API can be used to augment the current workflow.
2. If yes, select one API to augment the current workflow.
3. Decide whether to append the selected API to an existing API-use chain or create a new API-use chain:
    3.1 **Append to existing chain**: Choose this when the API logically should follow another call in an existing chain. This includes cases where:
        - The `tool_input` contains data from a previous API's `response`
        - The API's purpose depends on results from a previous call
        - Even if `tool_input` is minimal, the API conceptually continues work from a previous step
        When you decide to append, you should also select which API-use chain to append to.
    3.2 **Create new chain**: Choose this when the API is independent and does not logically depend on any existing API execution. For example:
        - The API addresses a completely separate aspect of the workflow
        - The `tool_input` is self-contained and doesn't rely on previous results
\end{lstlisting}

\begin{lstlisting}[style=prompt, caption={``Inverse Prediction'' template.}, label={lst:inverse-prediction}]
You are generating training data for a tool-use language model. Given API execution traces, create a natural user query that would trigger these API calls, followed by an appropriate response.

**API Execution Details:**
{api_use_chains}

**Task:** Generate (1) a natural user query and (2) the agent's response based on the API execution traces above.

**Important:** You will receive the API execution chains for context, but you should NOT return them in your output. Only return the query and response fields.

---

**CRITICAL: User Query Requirements**

1. ** DO**: Write queries like a real human would ask
   - "What's the weather forecast for London next week?"
   - "I'm researching Tesla stock. Show me recent performance and news."
   - "Find me Italian restaurants near Golden Gate Park with good ratings."

2. ** NEVER**: Mention APIs, tools, functions, or technical implementation
   - Never say: "call the weather API", "use get_forecast", "invoke the search tool"
   - Never ask: "which API should I use", "can you run this function"
   - Never include: tool names, API endpoints, function signatures

3. ** DO**: Provide specific, concrete details
   - Include exact values from tool_input (locations, IDs, names, numbers)
   - Use specific examples: "123 Main St, Oakland CA" not "an address"
   - Mention precise entities: "Tesla stock" not "a company's stock"

4. ** DO**: Create realistic scenarios
   - Explain WHY the user needs this information:
     * "I'm planning a trip..."
     * "I'm writing a research report on..."
     * "I need to prepare for a meeting about..."
   - Make the request feel natural and purposeful

5. ** DO**: Cover ALL API calls implicitly
   - If 3 APIs were called, the query should naturally require all 3
   - Don't list them ("do A, B, and C"), weave them into a cohesive need
   - Example: Instead of "Get weather, find hotels, search restaurants"
     --> "I'm visiting Paris this weekend. What should I expect, and where should I stay and eat?"

---

**Response Requirements:**
- Synthesize all API execution results into a helpful, natural response
- Present information clearly without mentioning APIs or tools
- Reference concrete data from the execution outputs
- Sound like a knowledgeable assistant answering a user's question

---

**Examples:**

**Example 1:**
API Chains: [weather(city="Tokyo"), currency_convert(from="JPY", to="USD", amount=5000)]
Query: "I'm traveling to Tokyo next month. What's the current weather like, and how much is 5,000 yen in US dollars?"
Response: "Tokyo is currently experiencing mild temperatures around 18C with partly cloudy skies. As for the currency conversion, 5,000 Japanese yen is approximately 33 US dollars."

**Example 2:**
API Chains: [github_search(topic="ML"), github_get_repo(id="tensorflow/tensorflow"), github_get_contributors(id="tensorflow/tensorflow")]
Query: "I'm researching popular machine learning projects on GitHub. Can you tell me about TensorFlow -- how active is the project and who are the main contributors?"
Response: "TensorFlow is one of the most popular machine learning frameworks on GitHub with over 180,000 stars. The project is very active with regular updates. The main contributors include members of the Google Brain team, with key developers like Jeff Dean and Rajat Monga being significant contributors."

---

Make the query sound like something a real person would ask in a conversation or search bar.

\end{lstlisting}

\begin{lstlisting}[style=prompt, caption={LLM judge template.}, label={lst:llm-judge}]
Task Overview:
You are an expert evaluator for a Tool-Use agent. Your task is to determine if the provided "Tool use trace" successfully retrieves the information needed to satisfy the "User query". 

CRITICAL RULES:
1. STRICT EVALUATION: Evaluate ONLY the information contained in the "Tool use trace". Do NOT assume the agent can answer using external knowledge.
2. ZERO SCORE: If the "Tool use trace" is empty, or if ALL tool calls in the trace returned an error (e.g., timeout, invalid parameters, authentication failure), the score MUST be 0.
3. GROUNDING: We are judging the SUFFICIENCY of the retrieved data to answer the query, not a natural language response.

Evaluation Criteria:
1. Atomic Request Decomposition:
   - Breakdown the user query into distinct, atomic information needs (e.g., "Request 1: Find movie X", "Request 2: Get cast for movie X").

2. Individual Request Scoring:
   - For EACH atomic request, assign a score based on the sufficiency of retrieved data:
     - 0: No information retrieved. No tools called, or the trace is empty, or ALL relevant tool calls failed (e.g., timeouts, authentication errors). 
     - 25: Unsuccessful attempt. Tools were called with intent but returned errors (e.g., "invalid parameter", "missing argument"), and NO useful data was retrieved. This acknowledges the effort and the error message provides information for a next step improvement.
     - 50: Partial success. Some relevant data was retrieved, but it is incomplete or lacks critical details (e.g., found the movie ID but failed to get its streaming links).
     - 75: Near complete success. The bulk of the requested information is present. Minor details might be missing, or the retrieval was slightly indirect/cluttered but sufficient.
     - 100: Complete success. All requested information was retrieved accurately via valid, grounded tool calls with no errors.
   - You can choose scores between the above values to better fit the real scenario.

3. Final Score Calculation:
   - Compute the final score by averaging the scores of all identified atomic requests.

Example:
Query: "Find 'Documentary' on Vimeo AND get streaming link for YT ID '123'" (2 requests)
- Trace: Vimeo search succeeded. YT retrieval failed. -> Score: (100 + 0) / 2 = 50.
- Trace: Both tools failed with timeouts. -> Score: (0 + 0) / 2 = 0.
- Trace: No tools called. -> Score: 0.

Input Data:
User query:
{query}

Tool use trace (a list of objects with tool_name, tool_description, tool_input, and response):
{tool_use_trace}
\end{lstlisting}
\subsection{Tool-use Error Handling}
Real-world tools may lead to execution failure, such as network timeouts and invalid parameters.
In the ``API Executor'' module, we configure the timeout to 10 seconds.
When an ``API Executor'' fails, it will reflect in the corresponding reports (see ``API Execution Report'' in~\autoref{fig:pipeline}).
The followup ``API Selector'' will not consider those failure report and only perform selection on those APIs that lead to successful API calls.

Our system cannot self-instruct while generating the data.
That being said, ``API Executor'' cannot learn from the tool-use experiences of other ``API Executor'' within or even outside a generation session.
To further enhance our system, we encourage future work to incorporate a memory system in our current implementation of \systemname.

\subsection{API Library}
We used the API library provided by ToolBench~\cite{qin2023toolllmfacilitatinglargelanguage}.
We found some API names and their corresponding configuration are not well annotated (\eg, APIs named as ``test\_v5'', ``test\_for\_test'', etc.), which negatively affects our generation.
Therefore, we used \textit{gemini-2.5-flash-lite} to perform two rounds of filtering.
In the first round, we filter APIs with low-quality annotations. In the second round, we create an agent to execute the tools, and aim to receive a non-failure response at least once within 10 LLM call budgets.
See \autoref{lst:api-filter-agent} for our instruction for the data filtering agent.
This results in 15,368 qualified APIs.

\begin{lstlisting}[style=prompt, caption={Instruction for the ToolBench-API-filtering agent.}, label={lst:api-filter-agent}]
You are an expert at testing APIs. Your goal is to successfully call the given API at least once.
You will be provided with the API documentation.
You should generate valid JSON inputs for the API.
If the API call fails, analyze the error message and try a different valid input.
You have a maximum of 10 attempts.

IMPORTANT:
1. If you believe it is impossible to get a successful response (e.g., API is permanently broken, requires unreachable dependencies), you MUST set "action" to "stop".
2. Do NOT simply repeat the same input if it failed. You MUST try different parameters or values.
3. If you decide to "call" the API, you MUST provide "tool_input".

Output your response in the following JSON format:
{
  "thought": "Your reasoning for the current attempt.",
  "action": "call" or "stop",
  "tool_input": { ... key-value pairs for the API arguments, required if action is 'call' ... }
}
\end{lstlisting}

\section{Model Training}

\subsection{Configurations}
\label{appendix: train_config}
The training code is based on "SFTTrainer" in the "trl" package.
We trained ToolGrad-1B based on "gemma-3-1b-it" using learning rate of 1e-05.
We trained ToolGrad-4B and ToolGrad-12B based on "gemma-3-4b-it" and "gemma-3-12b-it" using LoRA adapters with a rank of 64, an alpha of 16, and a dropout rate of 0.1.
The learning rate is 5e-06 and 2e-05, respectively.
All ToolGrad models are trained with batch size of 1, using the adamw optimizer for three epochs. For ToolBench models, we followed the official configuration reported in the paper.
All models are trained with 8k context windows size, and we computed the training loss exclusively on the assistant's completions by masking the user instruction tokens, i.e., "assistant\_only\_loss=True" in "SFTConfig".

\subsection{Training Budgets}
\label{appendix:training_budgets}
ToolGrad-1B, 4B, 12B was trained on four A100 GPUs and costs 1, 1.67, and 2.67 GPU hours in total, respectively. ToolBench-1B, 4B, 12B was trained on eight H100 GPUs and cost 29, 68, and 370 GPU hours in total, respectively.

\section{Dataset Formatting}
\label{appendix: dataset}
\autoref{lst:sft_formater} shows how we setup the dataset for our SFT tasks.
In additional to the 1-on-1 formatting, we also add 20\% more negative samples where we replace all the positive tools in the selection pool with unrelated tools, and the corresponding output is ``[]'', aiming to reduce model hallucination.

We recommend the readers to review our attached dataset (under ``train.jsonl'')for examples of our generated data.

\begin{lstlisting}[style=prompt, caption={Instruction template for SFT.}, label={lst:sft_formater}]
You are an expert in composing functions. You are given a question and a set of possible functions. Based on the question, you will need to make one or more function/tool calls to achieve the purpose.
If none of the functions can be used, point it out. If the given question lacks the parameters required by the function, also point it out.
You should only return the function calls in your response.

If you decide to invoke any of the function(s), you MUST put it in the format of [func_name1(params_name1=params_value1, params_name2=params_value2...), func_name2(params)]
You SHOULD NOT include any other text in the response.

The following is a list of apis in the library that you can use.
Each api is a JSON object with 'name', 'description' and 'parameters'.

```json
{selection_pool}
```
\end{lstlisting}

\section{Evaluation}

\subsection{LLM Judges}
\label{appendix:llm-judge}
We utilized multiple LLMs as judges to evaluate the prediction, including two SoTA proprietary models (gemini-2.5-pro, claude-4.5-sonnet) and two SoTA opensourced models (deepseek-v3.2, qwen3-235b). 
\autoref{lst:llm-judge} shows the prompt template we used for our judges. This approach is inspired by~\cite{qin2023toolllmfacilitatinglargelanguage}'s evaluation design. We improve their approach by only showing the LLM judges the query and the tool use traces without a generated response.
Our design can effectively avoid biases introduced in the response writer LLM, where the write may answer a given user query even though there is no tool call.

\subsection{Supplementary Discussion on ToolBench Results}
\label{appendix:toolbench_discuss}
Table 3 did not contain the results of ToolBench models under the "standard" condition. The ToolBench models are trained to output texts in the format of ``Thought: ... Action: ... Action Input: ...'', and call one single tool in each round.
The "standard" condition allows only one round of LLM execution and expects the model to predict all tool use.
Therefore, the ToolBench model cannot be fairly evaluated in this single-turn tool-use condition, as it will make at most one tool call.
The format limitation also explains why all Gemma-3 models perform worse in the ReAct/DFS condition -- base Gemma models cannot strictly follow this customized template, so the ToolBench evaluation parser fails to parse tool calls.

This explains 1) why ToolBench models cannot be fairly evaluated in our single-turn tool use condition (``standard''), and 2) why all Gemma-3 performs worse in ReAct/DFS condition -- base Gemma models cannot strictly follow this customized template, and thus the ToolBench evaluation parser fails to parse tool calls.

\subsection{Human Evaluation Design}
\label{appendix:human-eval}
We recruited two volunteers to annotate the results. Both participants work full-time as programmers in the AI industry.
We obtained participants' consent before the study.
This study is determined exempt by an ethics review board.

Here we provide details of our human evaluation study design mentioned in our response to Weakness 2. For each question, one author manually breaks it down into a list of subqueries.
The following query is an example we selected in the evaluation:

\myquote{I want to explore different genres of movies. Fetch the genre names and IDs for me. Also, provide me with the basic information about a specific cast member, including their name, profession, birth and death years, and best titles.}

For this query, we break it down into two subqueries:
\begin{enumerate}
    \item \textit{“Fetch the genre names and IDs for me.”},
    \item \textit{“Also, provide me with the basic information about a specific cast member, including their name, profession, birth and death years, and best titles.”}
\end{enumerate}

For each subquery, the rater needs to choose one score from 0, 25, 50, 75, 100. The definition is the same as the following quote in Prompt 5:

\begin{itemize}
    \item[0:] \textit{No information retrieved}. No tools called, or the trace is empty, or ALL relevant tool calls failed (e.g., timeouts, authentication errors).
    \item[25:] \textit{Unsuccessful attempt}. Tools were called with intent but returned errors (e.g., "invalid parameter", "missing argument"), and NO useful data was retrieved. This acknowledges the effort, and the error message provides information for the next step improvement.
    \item[50:] \textit{Partial success}. Some relevant data was retrieved, but it is incomplete or lacks critical details (e.g., found the movie ID but failed to get its streaming links).
    \item[75:] \textit{Near complete success}. The bulk of the requested information is present. Minor details might be missing, or the retrieval was slightly indirect/cluttered but sufficient. 
    \item[100:] \textit{Complete success}. All requested information was retrieved accurately via valid, grounded tool calls with no errors.
\end{itemize}

The final human evaluation score for a given answer to the full query is the average across all subqueries. This definition aligns with those in Prompt 5 to ensure we follow the same instructions in both human and LLM studies.

\subsection{BFCL Subscale Results}

BFCL evaluates tool-use capability in different scenarios.
``Simple'' involves a single invocation of a provided tool.
``Multiple'' tests the model's ability to select the correct tool from a candidate set.
``Parallel'' evaluates the generation of multiple simultaneous calls for a single tool.
``Multiple Parallel'' assesses the combined ability to select from multiple tools and generate multiple parallel calls within a single turn
Most of our formatted data are ``Multiple Parallel'' data.

\begin{table*}[t]

\centering
\begin{tabular}{cccccccccc}
\toprule
\multicolumn{2}{c}{\multirow{2}{*}{Model}} & \multicolumn{4}{c}{Non-live} & \multicolumn{4}{c}{Live} \\
\cmidrule(lr){3-6} \cmidrule(lr){7-10}
\multicolumn{2}{c}{} & Simple & Multi & Par & MultiPar & Simple & Multi & Par & MultiPar \\
\midrule
\multirow{3}{*}{\begin{tabular}[c]{@{}c@{}}Gemini\\ 2.5\end{tabular}} & Pro & \underline{78.67} & \textbf{94.00} & \underline{93.50} & \underline{92.00} & \underline{85.66} & 74.36 & 87.50 & \textbf{83.33} \\
& Flash & 77.33 & 91.50 & \textbf{96.00} & 87.50 & \textbf{87.21} & 75.97 & 81.25 & \underline{75.00} \\
& Lite & 70.08 & 86.00 & 90.00 & 89.50 & 67.05 & 51.66 & 75.00 & 50.00\\
\midrule
\multirow{3}{*}{GPT-5} & Base & 72.50 & 86.50 & 82.00 & 77.50 & 77.52 & 67.43 & 68.75 & 58.33 \\
& Mini & 59.17 & 72.50 & 71.50 & 69.00 & 69.77 & 61.16 & 75.00 & 37.50 \\
& Nano & 69.25 & 86.00 & 87.50 & 80.50 & 76.36 & 69.71 & 68.75 & 54.17 \\
\midrule
\multirow{3}{*}{\begin{tabular}[c]{@{}c@{}}Claude\\ 4.5\end{tabular}} & Opus & \textbf{79.58} & 93.50 & 93.00 & \textbf{92.50} & 84.50 & 74.17 & \underline{81.25} & 62.50 \\
& Sonnet & 47.25 & 79.50 & 53.50 & 59.00 & 73.26 & 40.17 & 56.25 & 33.33 \\
& Haiku & 55.67 & 84.00 & 38.00 & 44.00 & 66.67 & 49.76 & 56.25 & 16.67 \\
\midrule
\multicolumn{2}{c}{Hammer-2.1-7B} & 72.50 & 92.50 & 91.00 & 86.00 & 66.67 & 69.99 & 75.00 & 75.00 \\
\midrule
\multicolumn{2}{c}{ToolACE-2-8B} & 73.42 & 91.00 & 93.00 & 91.00 & 71.32 & \textbf{79.39} & 68.75 & 62.50 \\
\midrule
\multirow{3}{*}{\begin{tabular}[c]{@{}c@{}}Gemma 3\end{tabular}} & 12B & 76.25 & \textbf{94.00} & 91.00 & 56.50 & 85.66 & 71.89 & \textbf{87.50} & 45.83\\
 & 4B & 64.50 & 88.00 & 56.00 & 36.00 & 70.93 & 59.35 & 25.00 & 41.67\\
 & 1B & 43.33 & 36.00 & 0.00 & 1.50 & 36.43 & 6.27 & 0.00 & 0.00\\
\midrule
\multirow{3}{*}{\begin{tabular}[c]{@{}c@{}}\textbf{ToolGrad}\end{tabular}} & 12B & 75.25$\downarrow$& \textbf{94.00}$\uparrow$& \underline{93.50}$\uparrow$& 88.50$\uparrow$& \underline{85.66}$\uparrow$& \underline{77.11}$\uparrow$& 75.00$\downarrow$& 62.50$\uparrow$ \\
 & 4B & 65.33$\uparrow$& 86.50$\downarrow$& 73.00$\uparrow$& 65.00$\uparrow$& 71.32$\uparrow$& 64.86$\uparrow$& 43.75$\uparrow$& 50.00$\uparrow$ \\
 & 1B & 49.08$\uparrow$& 33.50$\downarrow$& 34.00$\uparrow$& 21.00$\uparrow$& 30.62$\downarrow$& 9.97$\uparrow$& 6.25$\uparrow$& 4.17$\uparrow$ \\
\bottomrule
\end{tabular}
\caption{BFCL score breakdowns under Non-live and Live tracks. The test is divided in Simple (single call), Multiple (tool selection + call), Parallel (parallel calls), and Multiple Parallel (tool selection + parallel calls). We highlighted \textbf{the highest} and \underline{the second highest} scores in each breakdown scale. On the ToolGrad rows, we annotated whether the score has increased ($\uparrow$) or decrease ($\downarrow$) compared to its corresponding Gemma 3 models. ToolGrad models show increase ($\uparrow$) in 19 out of 24 subcales.}
\label{tab: bfcl_detailed_subscale}
\end{table*}

\begin{table}[t]
    \centering
    \footnotesize
    \setlength{\tabcolsep}{3.5pt} %
    \begin{tabular}{ll cccc}
      \toprule
      \multicolumn{2}{l}{\multirow{2}{*}{Model}} & Non-live & Live & \multicolumn{2}{c}{Halluc.} \\
      \cmidrule(lr){3-3} \cmidrule(lr){4-4} \cmidrule(lr){5-6}
      & & Overall & Overall & Rel. & Irrel. \\
      \midrule
      \multirow{3}{*}{\shortstack[l]{Gemini\\2.5}} & Pro & \underline{89.54} & 76.83 & 62.50 & 86.97 \\
      & Flash & 88.08 & \underline{78.16} & 91.09 & 62.50 \\
      & Lite & 83.90 & 54.85 & 93.33 & 50.00 \\
      \midrule
      \multirow{3}{*}{\shortstack[l]{GPT-5}} & Base & 79.62 & 69.21 & 73.42 & 87.10 \\
      & Mini & 68.04 & 62.55 & 55.71 & 93.75 \\
      & Nano & 80.81 & 70.69 & 45.75 & 93.75 \\
      \midrule
      \multirow{3}{*}{\shortstack[l]{Claude\\4.5}} & Opus & \textbf{89.65} & 76.02 & 90.75 & 68.75 \\
      & Sonnet & 59.81 & 46.56 & 95.03 & 37.50 \\
      & Haiku & 55.42 & 52.48 & 95.29 & 31.25 \\
      \midrule
      \multicolumn{2}{l}{Hammer-2.1-7B} & 85.50 & 69.50 & 50.00 & 90.12 \\
      \midrule
      \multicolumn{2}{l}{ToolACE-2-8B} & 87.10 & 77.42 & 75.00 & 90.79 \\
      \midrule
      \multirow{3}{*}{Gemma 3} & 12B & 79.44 & 74.24 & 70.29 & 93.75 \\
      & 4B & 61.12 & 60.84 & 53.94 & 100.00 \\
      & 1B & 20.21 & 11.84 & 33.18 & 37.50 \\
      \midrule
      \multirow{3}{*}{\textbf{ToolGrad}} & 12B & 87.81$\uparrow$ & \textbf{78.46}$\uparrow$ & 93.75 & 59.27 \\
      & 4B & 72.46$\uparrow$ & 65.58$\uparrow$ & 93.75 & 59.27 \\
      & 1B & 34.40$\uparrow$ & 13.77$\uparrow$ & 81.25 & 26.89 \\
      \bottomrule
    \end{tabular}
    \caption{BFCL subscale scores. We highlight \textbf{highest} and \underline{second highest} scores. $\uparrow$ indicates increase over baseline Gemma 3 models. Halluc. (Hallucination) is split into Rel. (Relevance) and Irrel. (Irrelevance).}
    \label{tab: bfcl_subscale_1}
\end{table}

\section{Three Frameworks: Standard, ReAct and DFS}
\label{appendix: three_frameworks}
\begin{figure}
    \centering
    \includegraphics[width=\linewidth]{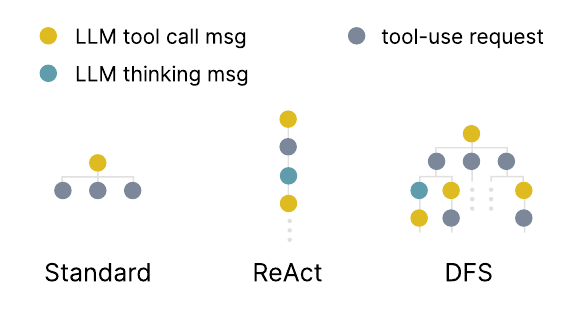}
    \caption{A visualized comparison among standard, ReAct, DFS inference frameworks.
    }
    \label{fig:three_frameworks}
\end{figure}

This work involves three different inference frameworks: 1) standard (\ie, the \systemname inference framework), 2) ReAct, 3) DFS.
\autoref{fig:three_frameworks} visualizes their differences.
In the standard framework that \systemname models use, an LLM is trained to predict multiple tool call requests in one shot, and thus, there is one single LLM step in the inference time.
On the other hand, ToolLlama models~\cite{qin2023toolllmfacilitatinglargelanguage} are trained to incorporate the ReAct~\cite{yao2023react} (a.k.a, CoT~\cite{wei2022chain}) framework.
In the ReAct framework, each LLM step returns a single tool call request.
The LLM and tool use is called alternatively, with an optional thinking step inserted in between.
The DFS framework extends the ReAct concept by enabling a tree search.

\section{License For Artifacts}
This work has used ToolBench for data generation and benchmark.
ToolBench is licensed under the Apache License 2.0, so we argue that our use is considered a fair use of the artifact.

Additionally, we will also open-source our source code, dataset, and fine-tuned models. These artifacts will be under the Apache License 2.0.

\section{AI Assistant Usage}
We utilized an AI assistant to support grammatical checks and code writing in this project.

\end{document}